# BomJi at SemEval-2018 Task 10: Combining Vector-, Pattern- and Graph-based Information to Identify Discriminative Attributes

**Enrico Santus[1], Chris Biemann[2], Emmanuele Chersoni[3]**
esantus@mit.edu,
biemann@informatik.uni-hamburg.de,
emmanuelechersoni@gmail.com
[1] Massachussetts Institute of Technology,
[2] Universität Hamburg,
[3] Aix-Marseille University

## Abstract

This paper describes BomJi, a supervised system for capturing discriminative attributes in word pairs (e.g. *yellow* as discriminative for *banana* over *watermelon*). The system relies on an XGB classifier trained on carefully engineered graph-, pattern- and word embedding-based features. It participated in the SemEval-2018 Task 10 on Capturing Discriminative Attributes, achieving an F1 score of 0.73 and ranking 2nd out of 26 participant systems.

## 1 Introduction

The recent introduction of popular software packages for training *neural word embeddings* (Mikolov et al., 2013a,b; Levy and Goldberg, 2014) has led to an increase of the number of studies dedicated to lexical similarity and to remarkable performance improvements on related tasks (Baroni et al., 2014).

However, the validity of similarity estimation as the only benchmark for semantic representations has been questioned, for several reasons. One for all, most evaluation datasets provide human-elicited similarity scores, with the consequences that the ratings are subjective and the performance of some automated systems is already above the upper bound of the inter-annotator agreement (Batchkarov et al., 2016; Faruqui et al., 2016; Santus et al., 2016a).

Originally proposed as an alternative benchmark for Distributional Semantic Models (DSMs), the Discriminative Attributes task focuses instead on the extraction of *semantic differences* between lexical meanings (Krebs and Paperno, 2016): given two words and an *attribute* (i.e., a discrete semantic feature), a system has to predict whether the attribute describes a difference between the corresponding concepts or not (e.g. *wing* is an attribute of *plane*, but not of *helicopter*).

Since even related words may differ for some non-shared attributes (e.g. hypernyms and hyponyms), the ability of automatically recognize discriminative features would be an extremely useful addition for the creation of ontologies and other types of lexical resources and would make machine decisions interpretable, enabling human validation (Biemann et al., 2018). Moreover, one can think to applications to many other NLP domains, such as machine translation and dialogue systems (Krebs and Paperno, 2016).

In the present contribution, we describe the BomJi classification system, which we used for the identification of discriminative features between concept pairs. According to the official evaluation results provided by the organizers[1], our system ranked second out of 26 participants. Our score, $F1 = 0.73$ lags slightly behind the best score of $0.75$. After the evaluation period, we run further experiments including all investigated features and found that the system can achieve up to 0.75 F1 score.

## 2 Capturing Discriminative Attributes

### 2.1 Task and Dataset Description

The task of capturing discriminative attributes between words can be described as a binary classification task, in which the system has to assign a positive label if the feature is discriminative for the first concept over the second one, and a negative label otherwise.

In the test data, the first two words correspond to the concepts being compared (they are called, respectively, the *pivot* and the *comparison* term) and the third word is the feature, which could describe a discriminative attribute or not (some examples are shown in Table 1). In the paper, we will refer

---
[1] https://competitions.codalab.org/competitions/17326#results

| Pivot | Comparison | Feature | Label |
|---|---|---|---|
| belt | plate | buckles | 1 |
| orange | cherry | sections | 1 |
| razor | brush | mink | 0 |
| necklace | bracelet | clasp | 0 |

Table 1: Examples of triples from the training dataset.

| Dataset | Examples | Features | Split P-N |
|---|---|---|---|
| Training | 17,782 | 1,292 | 37.06%-62.94% |
| Validation | 2,722 | 576 | 50.1%-49.9% |
| Test | 2,340 | 577 | 44.74%-55.26% |

Table 2: Number of examples, distinctive features and Positive-Negative split for each dataset.

to the elements of the triples as $w1$, $w2$ and $feat$.

A training and validation set have been provided for system development (figures in Table 2).

### 2.2 Embeddings and Graphs

For the Discriminative Attributes task, we combined word embeddings, patterns and information extracted from a graph-based distributional model.

Concerning the word embeddings, we used the vectors produced by two popular frameworks for word embeddings: Word2Vec (Mikolov et al., 2013a,b) and GloVe (Pennington et al., 2014).[2] The Word2Vec Skip-Gram architecture is a single-layer neural network, based on the dot-product between word vectors, in which the vector representation is optimized to predict the context of a target word given the word itself. The context generally consists of a word window of a fixed width around the target. The other framework, GloVe, is similar to traditional count models based on matrix factorization (Turney and Pantel, 2010; Baroni et al., 2014), in the sense that vectors are trained on global word-word co-occurrence counts. In the case of GloVe, the training objective is to learn word vectors such that their dot product equals the logarithm of the probability of the word to co-occur (Pennington et al., 2014).

As for graph-based information, we used the JoBimText architecture introduced by Biemann and Riedl (2013). In JoBimText, lexical items are represented as the set of their $p$ most salient contexts, where the contexts are words connected to the target by a given syntactic link or by a lexical pattern, and saliency is defined as an association measure between target and context, such as Positive Local Mutual Information (Evert, 2004). Differently from vector models, similarity between words in JoBimText is simply based on the overlap count of their common contexts.

Regarding patterns, first we extracted sentences where words and their features co-occur from a web-scale sentence-based index of English web (Panchenko et al., 2018) and then we extracted the patterns connecting our target words.

### 2.3 Methodology

The predictions submitted for the evaluation of Task 10 were obtained with a system that consists of a classifier, the Extreme Gradient Boosting (or XGBoost, Chen and Guestrin (2016)), trained on vectors aggregating carefully engineered graph-, pattern- and word embedding features.

In this section, we provide an overview of each feature type, leaving the discussion of their contribution to Section 3. The total of 55,026 features we used can be divided into five major groups.

**CO-OCCURRENCE.** Thirteen features related to word and word-feature frequency were calculated on the basis of the information extracted from a corpus of $3.2B$ words, corresponding to about 20% of the Common Crawl. For each word-feature combination (i.e. $w1 - feat$ and $w2 - feat$), we calculated: i) the co-occurrence count; ii) word count; iii) feature count; iv) Positive Pointwise Mutual Information (PPMI (Church and Hanks, 1990)) between each word and the feature; v) Positive Local Mutual Information (PLMI (Evert, 2004)) between each word and the feature. Further, we added three features representing the subtractions between the values of i), iv) and v) for the two word-feature combinations.

**JOBIMTEXT.** Another set of twenty-four features comes directly from JoBimText. They were calculated after extracting information through the public accessible JoBimText API[3], which returns a JSON file containing - for every target - a sorted list of $N$ features and their association scores (up to $N = 1,000$). As JoBimText distinguishes features according to their POS and dependency roles (i.e. features are in form WORD#POS#DEPENDENCY), a given $feat$ may appear multiple times in different POS-dependency combinations. However, we found that $feat$ rarely appears in the top $N$ features of $w1$ and $w2$, so we calculated our features not only for the given targets (i.e. $w_x$), but also for the first among their top 10 neighbors for which $feat$ was found (i.e. $top(neighbor(w_x) \ni feat, max = 10)$) and the first among the top 10 $feat$ neighbors for which the target was found (i.e. $top(neighbor(feat) \ni w_x, max = 10)$). This allowed us to check also whether the neighbors of the given words were associated with the candidate discriminative attributes or *vice versa*. The features are defined as follows (here they are described only with reference to the query on $w_x$, but this should be generalized to the other cases):

- *prediction by rank*: it is 1 if $feat$ is ranked higher for $w_1$ than for $w_2$, 0 otherwise;

- *prediction by score*: it is 1 if the total score between $w1 - feat$ is higher than for $w_2 - feat$, 0 otherwise;

- *total score*: sum of the scores of $w_1 - feat$ if prediction by score is 1, of $w_2 - feat$ otherwise;

- *top rank*: top rank of $feat$ for $w_1$ if prediction by score is 1, for $w_2$ otherwise;

- *bottom rank*: last rank of $feat$ for $w_1$ if prediction by score is 1, for $w_2$ otherwise;

- *number of occurrences*: count of how many times a feature appears among the features of $w_1$ if prediction by score is 1, otherwise the occurrences among the features $w_2$ are counted;

- *which neighbor?*: integer showing whether the query was performed on $w1/w2$ (in this case it would be initialized to 0), or on its neighbors (in this case it would be initialized with the rank of the first neighbor where $feat$ was found);

- *which $feat$ neighbor?*: integer showing whether the query was performed on $w1/w2$ (in this case it would be initialized to 0) or on the $feat$ neighbors (in this case it would be initialized with the rank of the first $feat$ neighbor where $w1$ or $w2$ was found).

---

[2] The pre-trained vectors are available, respectively, at https://code.google.com/archive/p/word2vec/ (Google News, 300 dimensions) and at https://nlp.stanford.edu/projects/glove/ (Common Crawl, 840B tokens, 300 dimensions).

[3] See www.jobimtext.org

**WORD EMBEDDING FEATURES.** Mikolov et al. (2013a) showed how vector offsets encode semantic information. We decided to include five features computed from either the Word2Vec or the Glove vectors, in order to take advantage of the offset information.

They are computed, respectively, as: $cos((w1 - w2), feat)$, $cos((w1 - feat), (w2 - feat))$, $cos((w1 - feat), w2)$, $cos((w2 - feat), w1)$. Finally, also the cosine between the word vectors (i.e. $cos((w1, w2))$) has been included.

**WORD EMBEDDING VECTORS.** These features are the simple concatenation of the three vectors of $w1$, $w2$ and $feat$. Again, we have two versions of these features, one for Word2Vec and one for Glove.

**PATTERNS.** In order to characterize the relation between words and features, we used an index to extract patterns occurring between them, independently of the order in which they appeared, and limited the maximum number of results to 10,000 sentences.

The patterns consist of sequences of either lemmatized tokens, POS or dependency tags, which are used to abstract from the surface form, thereby increasing the recall. Since the number of extracted patterns was far too high, we decided to use only patterns with a frequency higher than 100, obtaining a set of 53,136 items, using the observed pattern frequency per word pair as a predictor.

## 3 Experiments

### 3.1 Choosing the Training Set

During the practice phase, we noticed that the training set and the validation set show very different distributions. Running 5-fold cross validation experiments on either dataset, we obtained very high scores (sometimes close to $0.95$). However, such scores did not generalize to the other dataset, where they dropped to about $0.60$.

This was only partially due to lexical memorization (some lexemes were present in multiple triples of the same dataset, cf. Levy et al. (2015); Santus et al. (2016b)). In fact, investigating the frequency of the words in the triples, we found that, on average, in our index, the first and the second words, $w1$ and $w2$, were about four times more frequent in the validation than in the training set (respectively $4.7M$ and $5.4M$ versus $0.9M$ and $1M$); similarly, the third word (i.e. $feat$) was almost twice more frequent in the validation than in the training set (i.e. $3.9M$ versus $2.9M$). When the test set was made available, we could verify that its frequency distribution resembled the one in the validation set, with the first and second words respectively at $3.3M$ and $2M$, and the third at about $4.5M$ occurrences.

Given these differences, we have chosen to train our system only on the validation set, tuning the hyper-parameters by means of 5-fold cross validation. Because of its small size, we decided to train our second submission on a derived training set (henceforth New Validation), consisting of the 2,722 triples from the validation set plus 2,278 triples randomly extracted from the training, for a total of 5,000 samples. The use of different training data was the only difference between the two submissions.

|   | **Feature Type** | **# Feat** | **Training/Test** *17547 vs. 2340* | | **Validation/Test** *2722 vs 2340* | | **NewValidation/Test** *5000 vs 2340* | |
|---|---|---|---|---|---|---|---|---|
|   |   |   | **F1** | **F1++** | **F1** | **F1++** | **F1** | **F1++** |
| 1 | Co-occurrence | *13* | 0.68 | 0.68 | 0.72 | 0.72 | 0.72 | 0.72 |
| 2 | W2V Features | *5* | 0.55 | NA | 0.66 | NA | 0.63 | NA |
| 3 | W2V + Vectors | *905* | 0.57 | 0.68 (1 & 3) | 0.67 | 0.75 (1 & 3) | 0.66 | 0.73 (1 & 3) |
| 4 | Glove Features | *5* | 0.61 | NA | 0.66 | NA | 0.67 | NA |
| 5 | Glove + Vectors | *905* | 0.62 | 0.66 (1 & 5) | 0.68 | 0.74 (1 & 5) | 0.68 | 0.73 (1 & 5) |
| 6 | JoBim Features | *24* | 0.53 | 0.68 (1 & 6) <br> 0.67 (1, 3 & 6) <br> 0.66 (1, 5 & 6) | 0.62 | 0.74 (1 & 6) <br> **0.75 (1, 3 & 6)** <br> *0.74 (1, 5 & 6)* | 0.62 | 0.74 (1 & 6) <br> **0.75 (1, 3 & 6)** <br> *0.73 (1, 5 & 6)* |
| 7 | Patterns | *53176* | 0.56 | 0.67 (1, 3, 6 & 7) <br> 0.67 (1, 5, 6 & 7) <br> 0.68 (1, 3, 5, 6 & 7) | 0.52 | **0.75 (1, 3, 6 & 7)** <br> 0.74 (1, 5, 6 & 7) <br> 0.74 (1, 3, 5, 6 & 7) | 0.51 | 0.71 (1, 3, 6 & 7) <br> 0.69 (1, 5, 6 & 7) <br> 0.69 (1, 3, 5, 6 & 7) |

Table 3: Results both in absolute (F1) and in incremental terms (F1++: in brackets the features used to obtain the score) on the test set, organized by training set. In bold, we report the best results. In bold-italics, we report the submitted systems.

### 3.2 Model Selection

During the practice phase, we performed experiments with several classifiers, including K-Neighbors (with $K = 3$), Decision Tree (with $max\_depth = 5$), Random Forest (with $max\_depth = 5$, $n\_estimators = 10$ and $max\_features = 1$), Multilayer Perceptron (with $alpha = 1$), AdaBoost and XGB (the latter two with default settings).

Before running the classifiers, we also used Linear Support Vector Classification (SVC) with $penalty =' l1'$ and we tested several values of $C$ (i.e. 0.05, 0.1, 0.25, 0.5, 1) for feature selection.

In almost all settings we found that the best performing classifiers were the Random Forest, the Multilayer Perceptron and, above all others, XGB. With respect to the value of $C$ for the feature selection, instead, we noticed that it varied in relation to the feature types, with minor influence on the performance of XGB (+/-2%). In the final submission, therefore, we opted for removing this step from the pipeline and for keeping the full vector.

Concerning feature selection, we found that the pattern features had a neutral effect on the performance during cross validation. Similarly we noticed that Glove and Word2Vec performed comparably. Thus, we opted for submitting the output of the systems without using the pattern features and only Glove features (Word2Vec had lower coverage on the dataset). As it can be noticed in Table 3, however, this decision has slightly lowered the performance of our system in the competition.

### 3.3 Feature Contribution

In order to measure the contribution of the features, we re-ran the experiments over the test set, after training our model on the three available sets: training, validation and new validation sets.

Results are reported in Table 3, in which it is easy to notice a few things: the performance is strongly related to the choice of the training set, with Validation being better that New Validation, which is in turn better than the original Training set; the thirteen co-occurrence features are those that provide the major contribution to the performance, reaching a F1 score of 0.72. Further useful features are the word embedding vectors (900 features), the word embedding features (5 features) and, to some extent, the information from JoBimText. Pattern-based features perform the worst, almost on par with random guessing.

The submitted systems do not correspond to the systems obtaining the best performance in post-evaluation experiments (see the bold and bold-italics scores in Table 3); this was due to the use of Glove instead of Word2Vec in our submitted systems, because none of the embedding models had an edge over the other in the validation process.

## 4 Conclusions

In this paper we have presented BomJi, a supervised system for capturing discriminative attributes in word pairs (e.g. *yellow* as discriminative for *banana* over *watermelon*). BomJi relies on an XGB classifier trained on carefully engineered

graph-, pattern- and word embedding-based features. In the paper we have reported the contribution for each features, discussing the model selection and showing that a major factor affecting the performance was the choice of the training data.

In the official Task 10 evaluation, our submitted systems achieved an F1 score of 0.73, ranking 2nd out of 26 participant systems.

# References

Marco Baroni, Georgiana Dinu, and Germán Kruszewski. 2014. Don't Count, Predict! A Systematic Comparison of Context-Counting vs. Context-Predicting Semantic Vectors. In *Proceedings of ACL*, pages 238–247.

Miroslav Batchkarov, Thomas Kober, Jeremy Reffin, Julie Weeds, and David Weir. 2016. A Critique of Word Similarity as a Method for Evaluating Distributional Semantic Models. In *Proceedings of the 1st Workshop on Evaluating Vector-Space Representations for NLP*, pages 7–12.

Chris Biemann, Stefano Faralli, Alexander Panchenko, and Simone Paolo Ponzetto. 2018. A Framework for Enriching Lexical Semantic Resources with Distributional Semantics. *Natural Language Engineering*, 24(2):265–312.

Chris Biemann and Martin Riedl. 2013. Text: Now in 2D! A Framework for Lexical Expansion with Contextual Similarity. *Journal of Language Modeling*, 1(1):55–95.

Tianqi Chen and Carlos Guestrin. 2016. XGBoost: A Scalable Tree Boosting System. In *Proceedings of the 22nd ACM Sigkdd International Conference on Knowledge Discovery and Data Mining*, pages 785–794.

Kenneth W. Church and Patrick Hanks. 1990. Word Association Norms, Mutual Information, and Lexicography. *Computational Linguistics*, 16(1):22–29.

Stefan Evert. 2004. *The Statistics of Word Cooccurrences: Word Pairs and Collocations*. Ph.D. thesis.

Manaal Faruqui, Yulia Tsvetkov, Pushpendre Rastogi, and Chris Dyer. 2016. Problems With Evaluation of Word Embeddings Using Word Similarity Tasks. In *Proceedings of the 1st Workshop on Evaluating Vector-Space Representations for NLP*, pages 30–35.

Alicia Krebs and Denis Paperno. 2016. Capturing Discriminative Attributes in a Distributional Space: Task Proposal. In *Proceedings of the 1st Workshop on Evaluating Vector-Space Representations for NLP*, pages 51–54.

Omer Levy and Yoav Goldberg. 2014. Neural Word Embedding as Implicit Matrix Factorization. In *Advances in Neural Information Processing Systems 27*, pages 2177–2185.

Omer Levy, Steffen Remus, Chris Biemann, and Ido Dagan. 2015. Do Supervised Distributional Methods Really Learn Lexical Inference Relations? In *Proceedings of NAACL-HLT*, pages 970–976.

Tomas Mikolov, Kai Chen, Greg Corrado, and Jeffrey Dean. 2013a. Efficient Estimation of Word Representations in Vector Space. *arXiv preprint arXiv:1301.3781*.

Tomas Mikolov, Ilya Sutskever, Kai Chen, Greg Corrado, and Jeffrey Dean. 2013b. Distributed Representations of Words and Phrases and Their Compositionality. In *Proceedings of the 26th International Conference on Neural Information Processing Systems - Volume 2*, NIPS, pages 3111–3119.

Alexander Panchenko, Eugen Ruppert, Stefano Faralli, Simone Paolo Ponzetto, and Chris Biemann. 2018. Building a Web-Scale Dependency-Parsed Corpus from Common Crawl. In *Proceedings of LREC*.

Jeffrey Pennington, Richard Socher, and Christopher Manning. 2014. Glove: Global vectors for word representation. In *Proceedings of EMNLP*, pages 1532–1543.

Enrico Santus, Emmanuele Chersoni, Alessandro Lenci, Chu-Ren Huang, and Philippe Blache. 2016a. Testing APSyn Against Vector Cosine on Similarity Estimation. In *Proceedings of PACLIC*, pages 229–238.

Enrico Santus, Alessandro Lenci, Tin-Shing Chiu, Qin Lu, and Chu-Ren Huang. 2016b. Nine Features in a Random Forest to Learn Taxonomical Semantic Relations. In *Proceedings of LREC*.

Peter D. Turney and Patrick Pantel. 2010. From Frequency to Meaning: Vector Space Models of Semantics. *Journal of Artificial Intelligence Research*, 37:141–188.